%% file: main.tex
\documentclass[conference]{IEEEtran}
\IEEEoverridecommandlockouts
\usepackage[ruled,vlined]{algorithm2e}
\usepackage{multirow}
\usepackage{multicol}
\usepackage{tabularx}
\usepackage{booktabs}
\usepackage{enumitem}
\usepackage{subfigure}
\usepackage{amsfonts}
\usepackage{amsmath}
\usepackage{pifont}
\usepackage{color}
\usepackage{marvosym}

\usepackage{cite}
\usepackage{amsmath,amssymb,amsfonts}
\usepackage{algorithmic}
\usepackage{graphicx}
\usepackage{textcomp}
\usepackage{xcolor}
\def\BibTeX{{\rm B\kern-.05em{\sc i\kern-.025em b}\kern-.08em
    T\kern-.1667em\lower.7ex\hbox{E}\kern-.125emX}}
\begin{document}

\title{Fact-Preserved Personalized News Headline Generation}

\author{
  \IEEEauthorblockN{
    Zhao Yang\IEEEauthorrefmark{1}\IEEEauthorrefmark{2}\IEEEauthorrefmark{3}, 
    Junhong Lian\IEEEauthorrefmark{1}\IEEEauthorrefmark{2}\IEEEauthorrefmark{3}, 
    Xiang Ao\textsuperscript{\Letter}\IEEEauthorrefmark{2}\IEEEauthorrefmark{3}\IEEEauthorrefmark{4}
    \thanks{*Both authors contributed equally to this work.}
    \thanks{\textsuperscript{\Letter}Xiang Ao is the corresponding author.}
  }
  \IEEEauthorblockA{
    \IEEEauthorrefmark{2} \textit{Key Lab of Intelligent Information Processing of Chinese Academy of Sciences (CAS),} \\
    \textit{Institute of Computing Technology, CAS, Beijing 100190, China.}
  }
  \IEEEauthorblockA{
    \IEEEauthorrefmark{3} \textit{University of Chinese Academy of Sciences, Beijing 100049, China.}
  }
  \IEEEauthorblockA{
    \IEEEauthorrefmark{4} \textit{Institute of Intelligent Computing Technology, Suzhou, CAS.}
  }
  \IEEEauthorblockA{
    \{yangzhao20s, lianjunhong23s, aoxiang\}@ict.ac.cn
  }
}

% \author{
% \IEEEauthorblockN{Anonymous}
% }

\maketitle

\begin{abstract}
Personalized news headline generation, aiming at generating user-specific headlines based on readers' preferences, burgeons a recent flourishing research direction. 
Existing studies generally inject a user interest embedding into an encoder-decoder headline generator to make the output personalized, while the factual consistency of headlines is inadequate to be verified. 
In this paper, we propose a framework \underline{F}act-Preserved \underline{P}ersonalized News Headline \underline{G}eneration (short for FPG), to prompt a tradeoff between personalization and consistency.
In FPG, the similarity between the candidate news to be exposed and the historical clicked news is used to give different levels of attention to key facts in the candidate news, and the similarity scores help to learn a fact-aware global user embedding.  
Besides, an additional training procedure based on contrastive learning is devised to further enhance the factual consistency of generated headlines.
Extensive experiments conducted on a real-world benchmark PENS\footnote{https://msnews.github.io/pens.html} validate the superiority of FPG, especially on the tradeoff between personalization and factual consistency.
\end{abstract}


\input{body/intro}

\input{body/survey}
\input{body/preliminary}
\input{body/method}

\input{body/exp_set}
\input{body/experiment}

\input{body/conclusion}
\input{body/acknowledgment}

\bibliographystyle{IEEEtran}
\bibliography{sample-base}
\end{document}

%% file: body/intro.tex
\section{Introduction}
\begin{figure}[t]
    \centering
    \includegraphics[width=0.85\linewidth]{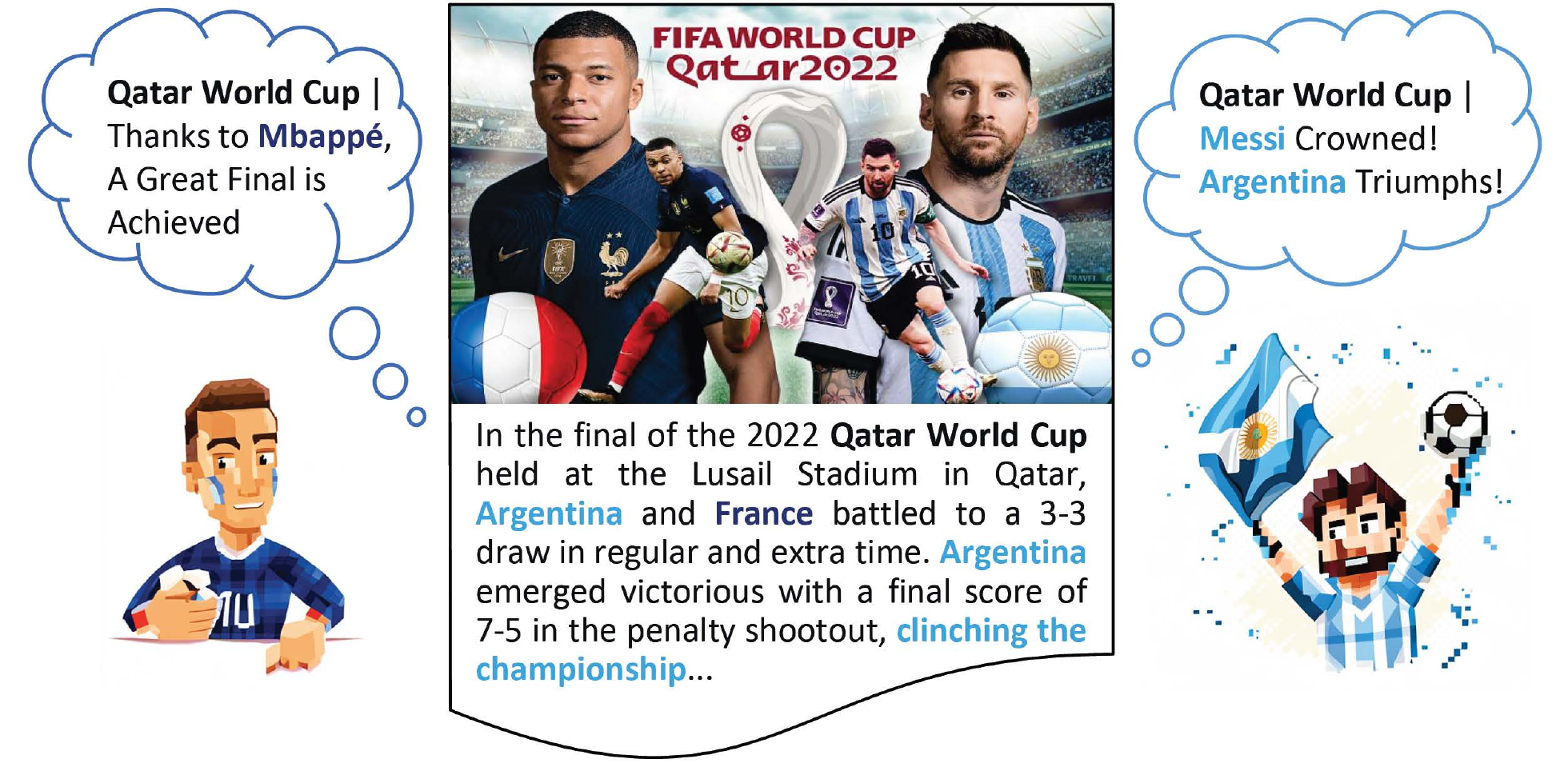}
    \caption{ An example illustrating the personalization in news headlines.}
\label{fig:intro:1}
\vspace{-2mm}
\end{figure}

News headline generation, intended to build a brief, informative, coherent headline for the given news article, has been perceived as a headline-specialized summarization task for decades~\cite{dorr2003hedge,alfonseca2013heady,lopyrev2015generating,takase2016neural,tan2017neural,luo2019reading,gavrilov2019self,gu2020generating,zhang2020pegasus,schick2021pretrain}.
Recently, personalized headline generation~\cite{ao2021pens}, i.e., generating a user-specific headline based on the user's reading interest, was proposed to produce eye-attracting headlines rather than potential clickbait.
Its underlying idea is that readers with different preferences can find their focal characters even in the same news, as illustrated in Fig.\ref{fig:intro:1}.
However, excessive personalization may threaten the factual consistency of news headlines, which is an imperative matter of principle in precision journalism~\cite{wagner2016framing}. 

To this end, we desire to reconcile the personalization and factual consistency of generated headlines.
The following challenges remain unsolved.
First, these two goals seem to run counter to each other. More personalization encourages more facts related to historical clicks in the headline, while high consistency requires preserving more facts from the candidate news in the title. 
Hence, jointly optimizing both goals in a unified framework might be challenging. 
Second, neither personalization nor factual consistency can be simply judged with existing metrics, a reasonable comprehensive evaluation method is in urgent demand.

To remedy these challenges, we propose a model named FPG~(\underline{F}act-Preserved \underline{P}ersonalized News Headline \underline{G}eneration), which utilizes an encoder-decoder framework that adapts Transformer~\cite{vaswani2017attention} with a history encoder, a personalized news encoder, and a user-guided decoder. 
The history encoder is analogous to existing work modeling users' interests based on their historical behaviors~\cite{wu2019neural,wu2019nrms,wu2019npa,an2019lstur}. 
The personalized news encoder leverages the similarity between the candidate news and historical clicks to attach various importance to clicked news.
The user-guided decoder learns a fact-aware user embedding to perturb headline generation based on personalized candidate news representations.
Furthermore, an enhanced training phase based on contrastive learning~\cite{khosla2020contrastive,gunel2021supervised} is leveraged for buoying the factual consistency of generated results. Similar techniques were recently observed effectively in abstractive summarization~\cite{nan2021quals}. 
For evaluation, we examine generated headlines based on personalization, factual consistency, and coverage, which will be detailed in Section~\ref{sec:expset-metric}.

In a nutshell, our contributions are: 
(1) We are the very first attempt to make a tradeoff between personalization and factual consistency for news headline generation. 
(2) We propose an end-to-end model FPG, equipped with a personalized news encoder that selectively concentrates on fact-consistent user interests via attention between the candidate news and historical clicks. 
Meanwhile, a training method based on contrastive learning takes factual consistency of the generation as a positive attribute. These two components are orthogonal to existing work.
(3) Extensive experiments on a real-world benchmark demonstrate the superiority of our model in generating fact-preserved personalized news headlines.

%% file: body/survey.tex
\section{Related Work}\label{sec:relatedwork}

Previous studies related to our task can be divided into two major categories: content-based headline generation and user-oriented headline generation.

\textbf{Content-based headline generation} aims to yield a concise, coherent, informative headline for the given article based on its content\footnote{
The content refers to information directly relevant to the article, including its domain, topic, category, etc.
}, which is similar to the text summarization task. 
The extractive approaches~\cite{dorr2003hedge,alfonseca2013heady} select a subset of actual sentences from the original article to compose a news summary, resulting in incoherent headlines with inadequate information.
The abstractive models~\cite{sun2015event,takase2016neural,tan2017neural,see2017get,gavrilov2019self} usually instantiate an encoder-decoder framework to build compact and coherent titles through learning the representations of the content.
In recent years, Transformer-based pre-trained models~\cite{devlin2019bert,raffel2020t5,lewis2020bart}
have reached SOTA for content-based headline generation~\cite{zhang2020pegasus,schick2021pretrain,liu-etal-2022-brio,li2022}.
However, these approaches have mediocre performance in personalized situation due to rare consideration for user preference.

\textbf{User-oriented headline generation} desires to build a headline that not only contains critical news facts but also grabs users' curiosity, promoting reading interests. 
This may require auxiliary user information, e.g., users' profile, landing page, historical clicks, etc. 
Some researchers propose to revamp headline styles~\cite{shen2017style} to attract readers' attention. Implicit approaches~\cite{shen2017style,fu2018style,prabhumoye2018style} differentiate the sentence into content and style representations to implicitly perform
style transfer. The explicit approaches~\cite{shu2018deep,zhang2018question,xu2019clickbait,Liu2022style} directly identify style-oriented examples or keywords for decorating titles. However, limited styles may not satisfy various users, and over-decorated headlines may also derive clickbait.

Recent studies on personalized text generation emphasize avoiding clickbait in engaging headlines~\cite{ao2021pens,xu2021review,wang2021reinforcing,zhang2022personalized}, but incorporating users' historical information may disrupt headline consistency due to global user embedding interference.

%% file: body/preliminary.tex
\section{Problem Formulation}\label{sec:preliminary}
The problem of personalized headline generation can be formulated as follows.
Given a user $u$, we denote $u$'s historical clicked news as $C_u=[c_1^u,c_2^u,\dots,c_{N}^u]$
, where $c_j^u$~($j=1,\dots,N$) is the $j$-th clicked news headline and $N$ is the length of the clicked sequence.
Each news headline $c$ is composed of a word sequence, i.e.,
$c=[w_1^c,w_2^c,\dots,w_T^c]$, 
where $T$ is the maximum length of the headline, $w_j^c \in \mathbb{V}$ for all
$1 \leq j \leq T$  and $\mathbb{V}$ is the word vocabulary.
Then, given a candidate news $v$ to be exposed to the user $u$ where its news body $X_v=[w_1^v,w_2^v,\dots, w_M^v]$ contains a maximum of $M$ words,
our target is to build a specific-customed headline $Y^u_v=[y^{u}_1, y^{u}_2, \dots, y^{u}_T]$
for the user $u$ based on his/her historical clicks, i.e., $C_u$, and the news body of $v$, i.e., $X_v$, where $y^{u}_j \in \mathbb{V}$ for all $1 \leq j \leq T$.

%% file: body/method.tex
\section{Methodology}\label{sec:method}
\begin{figure}[tbp]
\centering
\includegraphics[width=0.85\linewidth]{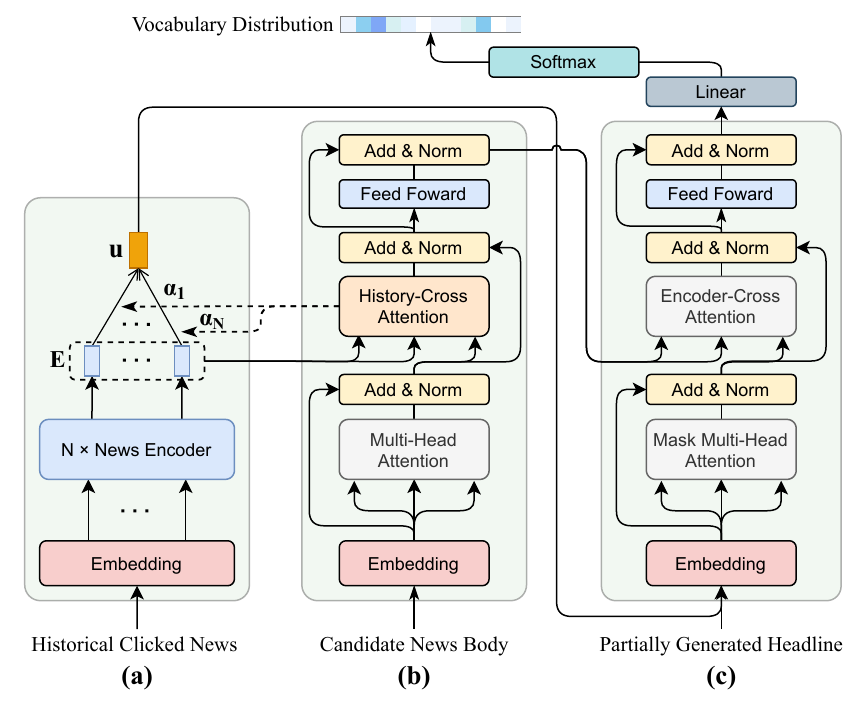}
\caption{\label{fig:model}
The framework of FPG. 
It has \textsf{N} layers of transformer blocks in both news encoder and decoder. ($\mathbf{a}$) is history encoder, ($\mathbf{b}$) is personalized news encoder, and ($\mathbf{c}$) is user-guided decoder.
}
\vspace{-2mm}
\end{figure}

This section details our proposed FPG model, which is illustrated in Figure~\ref{fig:model}, and we adopt Transformer~\cite{vaswani2017attention} as the backbone of FPG.

\subsection{History Encoder}
As demonstrated in Fig.\ref{fig:model}(a), the history encoder aims to learn users' interest representations based on their historical behaviors.
For each headline $c$ in the clicked sequence $C_u=[c_1^u,c_2^u,\dots,c_{N}^u]$ of the user $u$, the encoder first converts $c$ from a sequence of words into a sequence of embedding vectors, i.e., $\textbf{c} = [\textbf{{w}}_1^c,\textbf{{w}}_2^c,\dots,\textbf{{w}}_T^c]$,
$\textbf{{w}}_j\in{\mathbb{R}^{1\times {d_e}}}$.
Then, the embeddings are fed into a GRU~\cite{cho2014GRU} to learn the semantic hidden state of each word, i.e., $\mathbf{h}=[\mathbf{h}_1^c,\mathbf{h}_2^c,\dots,\mathbf{h}_T^c]$, $\mathbf{h}_j^c\in{\mathbb{R}^{1\times {d_e}}}$.
The weighted sum of $\mathbf{h}$ by Eq.(\ref{pre:news-level}) is considered the news representation of $c$.
\begin{equation}\label{pre:att1}\small
a_{j} = \mathsf{Softmax}(\mathbf{h}_j^c\mathsf{tanh}(\mathbf{V}_{a}
{\mathbf{h}_{j}^{c}}^\top+ \mathbf{b}_a))
\end{equation}
\begin{equation}\label{pre:news-level}\small
\mathbf{e_c} = \sum_{j=1}^{T} {a_j}\mathbf{h}_j^c 
\end{equation}
Where $\mathbf{V}_{a}\in{\mathbb{R}^{{d_e}\times {d_e}}}, \mathbf{b}_{a}\in{\mathbb{R}^{{d_e}\times 1}}$.
We denote $\mathbf{E}_u=[ \mathbf{e}_1,\mathbf{e}_2,\dots,\mathbf{e}_{N}]$ as the \textbf{news-level user interests} of $u$, where each $\mathbf{e}_{j}$ is obtained from the $j$-{th} news headline in $u$'s clicked sequence, i.e., $C_u$.

\subsection{Personalized News Encoder}
As shown in Fig.\ref{fig:model}(b), the personalized news encoder intends to encode a candidate news body based on the similarity between the candidate news and news-level interests of the corresponding user.
We expect the news body to exploit some valuable information from news-level user interests, which should share semantical similarity with partial content, to learn personalized representations.
Therefore, another \emph{history-cross attention} sub-layer is used to capture the interaction between news body and historical behaviors:
the query $\mathbf{Q}_h$ is the linear projection of the news body representations $\mathbf{X}$ while the key $\mathbf{K}_h$ and value $\mathbf{V}_h$ are projections of news-level user interest embeddings $\mathbf{E_u}$.
\begin{equation}\label{pe:q,k,v}\small
\mathbf{Q}_h = \mathbf{X}^\top\mathbf{H}^{Q}, \ 
\mathbf{K}_h = \mathbf{E}_u^\top\mathbf{H}^{K},\
\mathbf{V}_h = \mathbf{E}_u^\top\mathbf{H}^{V}
\end{equation}
\begin{equation}\label{pe:his-cross-att}\small
\mathbf{X}_{p}= \mathsf{Softmax}(\frac{\mathbf{Q}_h^\top \mathbf{K}_h}{\sqrt{d_e}}) \mathbf{V}_h
\end{equation}
Where $\mathbf{H}^{Q},\mathbf{H}^{K},\mathbf{H}^{V}\in \mathbb{R}^{d_e \times d_e}$ are learnable parameter matrices.
Through such interaction, information from historical clicks, which is semantically similar to the candidate news, is attached to the representations of the news body implicitly, enhancing attention to the user's fine-grained interests.
For example, analogous entities that appear both in clicked news and the candidate news directly reflect the user's potential interests should be spotlighted. 
After utilizing \textsf{N} encoder blocks, we obtain the history-aware representations of the candidate news, i.e., $\mathbf{X}^p_{enc}$.

\subsection{User-guided Decoder}
As illustrated in Fig.~\ref{fig:model}(c), the user-guided decoder generates a personalized headline under the guidance of a global user interest embedding.

Instead of learning a fixed embedding for each user~\cite{ao2021pens}, which may contain inconsistent information with the candidate news, our approach desires to learn a fact-aware global user representation based on the relevance of news-level interests to the candidate news. 
The user embedding is the weighted summation of news-level user interests:
\begin{equation}\label{ugd:user}\small
\mathbf{u} = \sum_{j=1}^{N} {{\alpha}_j\mathbf{e}_j}
\end{equation}
Where ${\{\alpha}_{1,\dots,N}\}$ are attention scores of history-cross attention sub-layer in the first news encoder block.

To avoid additional edits to the decoder input format or extra training parameters, we simply replace the \textsf{[BOS]}\footnote{A special token representing the beginning of a sentence.} token with the user embedding $\mathbf{u}$ so that the model again 
considers the user's preference at every decoding step, enhancing the personalization of the generated headline.
At each decoding step $t$, the input embeddings of the partially generated headline is $\mathbf{Y}^u= [\mathbf{u};\mathbf{y}_1,\dots, \mathbf{y}_{t-1}]$
,where $\mathbf{u}, \mathbf{y}_j\in{\mathbb{R}^{1\times d_e}}$, for all  $1 \leq j \leq {(t-1)}$.
$\mathbf{Y}^u$ is then fed into the masked self-attention layer and aligned with personalized encoder representations $\mathbf{X}^p_{enc}$. After \textsf{N} blocks, the output of the decoder at time step $t$ is $\mathbf{S} _t^{\mathsf{N}} \in \mathbb{R}^{1 \times d_e}$. 
The probability distribution $P$ over the whole vocabulary can be calculated as:
\begin{equation}\label{upd:dist}\small
P({\hat{y_{t}}}) = \mathsf{Softmax}({\mathbf{S} _t^{\mathsf{N}}}\mathbf{W}_{v}+\mathbf{b} _{v})
\end{equation}
Where $\mathbf{W}_{v}\in \mathbb{R}^{{d_e} \times \|\mathbb{V} \|}$ and $\mathbf{b} _{v} \in \mathbb{R}^{1 \times \|\mathbb{V} \|}$ are learnable parameter matrices. We use the
negative log-likelihood as the loss function to train the
headline generation model:
\begin{equation}\label{upd:nll}\small
\mathcal{L}_\mathit{NLL} = - \sum_{i=1}^{T}{ \log P({y}_i|{y}_1, \dots, {y}_{i-1};X,C )} 
\end{equation}
Where $T$ is the length of the generated headline.

\subsection{Fact-enhanced Training}
The modules mentioned above allow news-level and global user representations to be involved in personalized headline generation, while auxiliary user information may also bring inconsistency in headline generation. 
Especially when none of historical clicks are relevant to the candidate news, the user embedding may induce misinformation at the decoding step. Therefore, an additional mechanism is required to enhance the factual consistency of generated personalized headlines.

Previous studies have shown that simply removing unfaithful instances from the supervision data~\cite{matsumaru2020improving} or utilizing methods such as reinforcement learning~\cite{ijcai2020rl} and contrastive learning~\cite{cao2021cliff,nan2021quals} can enhance the consistency in text generation.
Motivated by~\cite{nan2021quals}, we apply a multi-stage fact-enhanced training phase, as demonstrated in Algorithm~\ref{alg:train}, to improve the factual consistency of generated headlines by minimizing a contrastive learning loss: 
\begin{equation}\label{ts:cll}\small
\begin{split}
\mathcal{L}_\mathit{CLL} = &- \underset{L_{C}^+}{\underbrace{{\mathbb{E}}_{x,c,y^+ \in \mathcal{D}^*}  {\log P({y}^{+};x,c )}}} \\
&-\underset{L_{C}^-}{\underbrace{{\mathbb{E}}_{x,c,y^- \in \mathcal{D}^*} {\log (1-P({y}^{-};x,c ))}}}  
\end{split}
\end{equation}

Training examples for contrastive learning (notated as $\mathcal{D}^*=\{ X, C, Y^+, Y^- \}$) are constructed from the news corpus.
We selected the prominently ranked headline samples with high factual accuracy scores compared to the news articles as positive instances. Additionally, we generated negative instances by deliberately designing positive instances with factual errors using a series of rule-based methods.

\begin{algorithm}[tbp]
\small
    \SetAlgoLined
    \textbf{Input:} $\mathcal{C}=\{ X, Y \}$, $\mathcal{D}_l=\{ X, C, Y \}$, $\mathcal{D}^*=\{ X, C, Y^+, Y^- \}$\;
    Initialize Transformer parameters $\xi$ with BART-base\;
    Other parameters $\theta$ are randomly initialized\;
    \textbf{1. Pre-train the headline generator with MLE}\;
    \textbf{2. Froze $\xi$ to train the history encoder}\;
    \For{epoch=1:$epoch_{2}$}{
      Sample $\{ X_i, C_i, Y_i \}$ \ from $\mathcal{D}_l$\;
      Update $\theta$ via minimizing Eq.(\ref{upd:nll})\;
    }
    \textbf{3. Train all parameters of FPG}\;
    \For{epoch=1:$epoch_{3}$}{
      Sample $\{ X_i, C_i, Y_i \}$ \ from $\mathcal{D}_l$\;
      Update $\theta$ and $\xi$ via minimizing Eq.(\ref{upd:nll})\;
    }
    \textbf{4. Fact-enhanced training}\;
     \For{epoch=1:$epoch_{4}$}{
      Sample $\{ X_i, C_i, Y_i^+, Y_i^- \}$ \ from $\mathcal{D}^*$\;
      Update $\xi$ via minimizing Eq.(\ref{ts:cll})\;
    }
\caption{Training schedule of FPG}
\label{alg:train}
\end{algorithm}

\begin{table}[htbp]
\renewcommand\tabcolsep{4.25pt}
\centering
\caption{
The statistics of datasets.  
$\mathcal{D}_T$ denotes the test set.
}
\label{table:statistics}
\begin{tabular}{c|cccccc}
    \toprule
    &$\mathcal{C}$ & $\mathcal{D}_{10}$ & $\mathcal{D}_5$ & $\mathcal{D}_3$ & $\mathcal{D}^*$ &$\mathcal{D}_T$\\ 
    \midrule
    \#news & 99,598 & 10,651 & 10,651 & 10,651 & 6,346 & 3,840 \\
    \#users & NA & 49,956 & 31,152 & 21,651 & 5,875 & 103 \\
    \#training examples & NA & 63,434 & 38,134 & 25,895 & 6,346 & 20,599 \\
    \bottomrule
\end{tabular}
% \vspace{-2mm}
\end{table}

%% file: body/exp_set.tex
\section{Experiment Settings}\label{sec:expsetting}

\subsection{Datasets Settings}\label{sec:expset-data}
We validate our proposed method on the PENS benchmark, which comprises a news corpus, $500,000$ anonymized user click behavior data from Microsoft News involving $445,765$ users, and manually annotated personalized headlines. The test set includes $50$ news of interest chosen by $103$ annotators to build their clickstream, along with $200$ news articles for which they provided preferred headlines, serving as personalized headlines. More details on PENS can be found in~\cite{ao2021pens}.

Due to the lack of reliable personalized headlines during the training phase, distant supervision is conducted to train our model. We take advantage of historical clicks to model a user's interests and approximate original headlines of newly clicked news within this impression as imperfect labels for training.
It's notable that considering some news that have appeared in the clickstreams of too many users as personalized headlines is unreasonable. 
To mitigate this problem, we have limited the number of users associated with each news during the training process. This limitation ensures that our model doesn't overly focus on news articles that have a broad appeal and have been clicked on by a vast number of users.
The training data with the limitation number $l$ is noted as $\mathcal{D}_l$. We use $\mathcal{D}_5$ for our major experiments, where the same news article in the training set is clicked by a maximum of five users.

In addition, we only pre-train the headline generator with the corpus that excludes candidate news used in the training and test set, indicated as $\mathcal{C}$. This decision stemmed from our observation that, despite achieving higher coverage scores, the model cannot acquire the capability to decorate user-specific headlines, which contradicts our goal of personalized headline generation.
The statistics of datasets are shown in Table~\ref{table:statistics}.

\subsection{Baselines}
Baselines consist of non-personalized and personalized approaches. We include some SOTA headline generation models:
(1) \textbf{PGN}~\cite{see2017get} is a seq2seq model with a copy mechanism.
(2) \textbf{PG+Transformer}~\cite{zhong2019searching} combines a transformer-based encoder with the pointer-generator network.
(3) \textbf{Transformer}~\cite{vaswani2017attention} is an encoder-decoder model based only on the attention mechanism.
(4) \textbf{BART} \cite{lewis2020bart} is a highly effective large pre-trained transformer-based model for text generation.
Besides, we also compare with some baselines mentioned in~\cite{ao2021pens}, including NPA~\cite{wu2019npa}, EBNR~\cite{okura2017embedding}, NRMS~\cite{wu2019nrms}, and NAML~\cite{wu2019neural}.

Our proposed model is denoted as \textbf{FPG-GRU}.
By replacing the GRU layer in our history encoder with other structures like CNN and Self-Attention layer, we have two more variants of FPG, including \textbf{FPG-CNN}, \textbf{FPG-SA}.

\begin{table*}[tbp]
\renewcommand\tabcolsep{15.0pt}
\centering\caption{The overall performances of compared methods. 
% The improvements of FPG over other baselines are significant with p-value $< 0.05$.
}
% \resizebox{\textwidth}{!}{
\begin{tabular}{c|cccccc}
% \midrule[1pt]
\toprule
\multirow{2}{*}{\textbf{Methods}} & \multicolumn{6}{c}{\textbf{Metrics}} \\
& $\mathbf{P_C}\textbf{(avg)}$ & $\mathbf{P_C}\textbf{(max)}$ & \textbf{FactCC} & \textbf{ROUGE-1} & \textbf{ROUGE-2} & \textbf{ROUGE-L}  \\ 
\midrule                                                            
PGN             & 2.71 & 16.20 & 65.08 & 19.86 & 4.76 & 18.83 \\    %1.81, 10.55
PG+Transformer  & 2.66 & 16.99 & 53.26 & 20.64 & 4.03 & 18.62 \\    % 1.91, 11.09
Transformer     & 2.70 & 16.36 & 61.61 & 19.54 & 4.72 & 16.36  \\   %1.69， 10.72
BART            & 2.72 & 17.13 & 86.67 & 26.27 & 9.88 & 22.85 \\    %1.42,9.37 
\midrule
PENS-NPA        & 3.43 & 21.48 & 51.51 & 21.08 & 4.03 & 19.03 \\ % 1.98, 12.17
PENS-EBNR       & 3.31 & 20.04 & 52.74 & 20.88 & 3.87 & 18.66 \\
PENS-NRMS       & 3.52 & 21.66 & 50.73 & 21.35 & 4.22 & 19.18 \\
PENS-NAML       & \textbf{3.93} & \textbf{22.73} & 50.16 & 22.81 & 4.90 & 19.65 \\     %1.62,10.61
\midrule
FPG-SA          & 2.89 & 17.20 & 89.20 & 27.13 & 10.39 & 23.09\\ % 1.51,9.41
FPG-CNN         & 2.88 & 17.20 & \textbf{89.29} & 27.20 & 10.41 & 23.15 \\ % 1.51, 9.40
FPG-GRU         & 2.88 & 17.27 & 89.26 & \textbf{27.33} & \textbf{10.51} & \textbf{23.30}\\ %1.50,9.46
\bottomrule
\end{tabular}
\label{tab:res}
\vspace{-2.5mm}
\end{table*}

\subsection{Evaluation Metrics}\label{sec:expset-metric}
Traditional metrics like ROUGE~\cite{lin2004rouge} mainly assess text-reference overlap and fail to capture headline personalization and consistency with content. Thus, we adopt a three-pronged approach to comprehensively evaluate headline quality.
\subsubsection{Personalization}
While lacking a valid metric for personalization, we can gauge it by comparing generated headlines to users' historically clicked titles, which reflect their fine-grained reading preferences as the personalization score.
\begin{equation}
\mathsf{P}_{sim}(\mathsf{max/avg}) = \underset{{c \ \in \ C_u}}{\mathsf{Max/Mean}}\ {sim}(c, y) 
\end{equation}
Where $C_u$ is the click sequence of user $u$, $y$ is the generated headline, ${sim}$ indicates similarity functions.
We report the mean and maximum value of all cosine similarity scores to evaluate fine-grained personalization, noted as $\mathsf{P}_C(\mathsf{max})$ and $\mathsf{P}_C(\mathsf{avg})$. 
A high maximum score indicates similarity to at least one reader's historically clicked title related to their interest, while the mean score reflects overall similarity between historical titles and the generated headline.

\subsubsection{Factual Consistency}
The factual consistency scores reflect the news headline's faithfulness to the source article.
We utilize FactCC~\cite{kryscinski2020factcc}, a weakly-supervised, model-based approach, to evaluate the factual consistency score. 

\subsubsection{Coverage}
We assess the informativeness and coverage of generated headlines by reporting the average F1 of ROUGE scores~\cite{lin2004rouge}. 
The coverage scores also partially reflect general personalization, given that manually-written headlines in test set mirror annotators' personalized reading preferences~\cite{ao2021pens}.

\subsection{Implementation Details}
The head number in multi-head attention layer is $12$. 
The number of encoder and decoder block \textsf{N} is $6$. 
The dimension $d_e$ is set to $768$.
All components of Transformer are initialized with BART-base parameters\footnote{https://huggingface.co/facebook/bart-base}.
The optimizer is AdamW~\cite{loshchilov2018adamw} with ${\beta}_1 = 0.9$ and ${\beta}_2 = 0.99$. 
The epoch number for the pre-trained phrase is $5$, and $5$, $9$, $1$ for each training stage afterward. The learning rates for each training stage are set to $3e-5$, $1e-4$, $3e-5$, and $1e-7$, respectively. 
During decoding, we use beam search with $beamsize=3$.
We trained and evaluated the model on a single NVIDIA V100 GPU.

%% file: body/experiment.tex
\section{Experiment Results}\label{sec:expresults}

\subsection{Performance Evaluation}
The main results are shown in Table~\ref{tab:res}. 
We evaluate generated personalized headlines through three aspects, namely coverage, factual consistency, and personalization.

\textbf{Coverage} indicates that our method FPG-GRU achieves the highest scores at ROUGE-1, -2, and -L with $27.33$, $10.51$, and $23.30$, significantly outperforming other baselines. This indicates that our model generates more informative, fluent headlines, and matches the users' general interests well.

\textbf{Factual consistency} issues were prevalent in earlier works and are more pronounced in current personalization methods, probably because emphasizing personalization compromises headline faithfulness.
We attribute BART's strong performance in factual consistency primarily to its ability to reconstruct the corrupted original text during the pre-training phase.

\textbf{Personalization} results reveal that personalized methods get higher personalization scores by modeling the user interests to inject personalized information, surpassing other non-personalized methods. Furthermore, it's worth noting that previous personalized models sometimes obtain higher personalization scores at the expense of headline consistency, potentially undermining the credibility of news. Our method is more like fact-preserving personalization, striking a balance between personalization and factual consistency in news headlines. While retaining BART's strong ability of consistency, we further enhance the user appeal of generated headlines.

\begin{table}[tbp]\small
\setlength{\abovecaptionskip}{3.0pt}
\setlength{\belowcaptionskip}{0.0pt}
    \centering
    \caption{A case of personalized news headlines from three models.}
    \label{table:case}
    \begin{tabularx}{0.48\textwidth}{X}
    \toprule
    \textbf{News Article} \\
    \underline{\textbf{Justin Rose}} didn't just dominate Thursday afternoon's marquee pairing at the 2019 U.S. Open, he \underline{\textbf{tied a record}} set by his more famous playing partner. With \underline{\textbf{an opening 65}}, Rose matched \underline{\textbf{Tiger Woods' first-round score in 2000}} for the lowest-ever \underline{\textbf{U.S. Open round at Pebble Beach}} \dots \\
    \midrule
    \textbf{Historical Clicks } \\
    \textcolor{red}{\textbf{US Open: Tiger Woods}} finishes strong at Pebble Beach\\
    \textcolor{red}{\textbf{Tiger Woods}} fought back Sunday and had his best \textcolor{red}{\textbf{U.S. Open}} score in 10 years\\
    2019 \textcolor{red}{\textbf{U.S. Open Tiger}} Tracker: \textcolor{red}{\textbf{Woods}} shoots second-round 72\\
    \midrule
    \textbf{Manually-written Headline}\\
    Justin Rose tied \textcolor{red}{\textbf{Woods}}' score of 65 in \textcolor{red}{\textbf{U.S. Open}} round at Pebble Beach \\
    \midrule
    \textbf{PENS-NAML}\\
    \textcolor{blue}{\textbf{Tiger Woods first round score} \ding{56}} U.S. Open \\
    \textbf{BART} \\
    Rose take the lead, ties \textcolor{red}{\textbf{Tiger' Pebble Beach record} \ding{52}} \\
    \textbf{FPG-GRU}\\
    Justin Rose ties \textcolor{red}{\textbf{Tiger Woods' U.S. Open record} \ding{52}} with opening 65 at Pebble Beach \\
    \bottomrule
    \end{tabularx}
    \vspace{-2mm}
\end{table}

\subsection{Case Study}
Finally, we exhibit an interesting case in our experiment, as shown in Table~\ref{table:case}. 
We compare our generated personalized headlines with outputs from the base headline generator, i.e., BART, and with personalized titles built by SOTA of personalized headlines generation, i.e., PENS-NAML.

Notably, previous models trained from scratch exhibited factual and syntactic errors. For instance, when the source news article reported ``Justin Rose'' as the golfer's score, the generated headline mistakenly mentioned ``Tiger Woods''. Analyzing the user's click history, it's evident he/she is a golf tournament enthusiast, possibly favoring ``Tiger Woods''. While personalized headlines should emphasize such interests, they must maintain factual consistency. In contrast, BART faithfully reflected the news content, generating a more coherent and factual headline despite lacking personalization. Meanwhile, our FPG-GRU strikes a balance between user appeal and factual consistency, offering a more personalized, informative, and consistent headline. It highlights relevant phrases like ``Tiger Woods'' and ``U.S. Open'' and provides additional details such as ``65'' and ``Pebble Beach'', aligning better with the user's interests.

%% file: body/conclusion.tex
\section{Conclusion and Discussion}\label{sec:conclusion}
In this paper, we proposed a framework FPG to make a trade-off between personalization and factual consistency in personalized news headline generation. 
This framework is underpinned by the principle of user appeal, leveraging the semantic similarity between the candidate news and the user's historical click patterns to selectively emphasize key facts that align with the user's nuanced interests.
Meanwhile, the global user embedding subtly influences the decoder's ultimate prediction, thereby infusing a degree of personalization into the generated headlines.
In the pursuit of consistency, we have engineered a fact-aware user embedding that serves to mitigate the propagation of inconsistent information. Additionally, we have implemented a contrastive learning-based factual enhancement training regimen, which bolsters the model's proficiency in preserving factual consistency between the generated headlines and the source news.
Extensive experiments conducted on the PENS benchmark demonstrate the superiority of our method over other baselines in the balance between personalization and fact-preservation.

Our focus on fact-preserving personalization makes generating high-quality personalized headlines particularly challenging when the candidate news lacks facts that align with the user's historical click pattern. These concerns have inspired us to advocate for further research to model the various interests of users, including innate preferences and behavioral tendencies, to generate more effective personalized headlines. 

%% file: body/acknowledgment.tex
\section*{Acknowledgment}

The research work supported by National Key R\&D Plan No. 2022YFC3303303, the National Natural Science Foundation of China under Grant (No.61976204). Xiang Ao is also supported by the Project of Youth Innovation Promotion Association CAS, Beijing Nova Program Z201100006820062.